\documentclass[10pt,twocolumn,letterpaper]{article}

\usepackage[pagenumbers]{cvpr} 

\definecolor{cvprblue}{rgb}{0.21,0.49,0.74}
\usepackage[pagebackref,breaklinks,colorlinks,allcolors=cvprblue]{hyperref}
\usepackage{hyperref}
\usepackage{url}
\usepackage{xcolor}
\usepackage{graphicx}
\usepackage{multirow}
\usepackage{bbding}
\usepackage{booktabs}

\usepackage{amsmath,amsfonts,bm}

\def\eqref#1{equation~\ref{#1}}

\def\1{\bm{1}}

\def\va{{\bm{a}}}

\def\vx{{\bm{x}}}

\def\mI{{\bm{I}}}

\DeclareMathAlphabet{\mathsfit}{\encodingdefault}{\sfdefault}{m}{sl}
\SetMathAlphabet{\mathsfit}{bold}{\encodingdefault}{\sfdefault}{bx}{n}

\def\confName{CVPR}
\def\confYear{2025}

\title{Diffusion Transformer Policy}

\author{Zhi Hou$^1$\footnotemark[1] \ \ \ \ Tianyi Zhang$^{2,1}$\footnotemark[1] \ \ \ \ Yuwen Xiong$^{1}$ \ \ \ \ Hengjun Pu$^{3,1}$ \ \ \ \ Chengyang Zhao$^{4,1}$ \\ 
Ronglei Tong$^{5}$ \ \ \ \ Yu Qiao$^{1}$ \ \ \ \ Jifeng Dai$^{6,1}$ \ \ \ \ Yuntao Chen$^7$\footnotemark[2] \\
$^1$ Shanghai AI Lab \ \ \ \
$^2$ 
College of Computer Science and Technology, Zhejiang University \\
$^3$ MMLab, The Chinese University of Hong Kong  \
$^4$ Peking University  \  \
$^5$ SenseTime Research \\
$^6$ Tsinghua University \ \ 
$^7$ Center for Artificial Intelligence and Robotics, HKISI, CAS \\
}

\begin{document}
\maketitle
\renewcommand{\thefootnote}{\fnsymbol{footnote}}
\footnotetext[1]{Equal Contribution}
\footnotetext[2]{Corresponding Author}
\begin{abstract}


Recent large vision-language-action models pretrained on diverse robot datasets have demonstrated the potential for generalizing to new environments with a few in-domain data. However, those approaches usually predict individual discretized or continuous action by a small action head, which limits the ability in handling diverse action spaces. 
In contrast, we model the continuous action sequence with a large multi-modal diffusion transformer, dubbed as Diffusion Transformer Policy, in which we directly denoise action chunks by a large transformer model rather than a small action head for action embedding. 
By leveraging the scaling capability of transformers, the proposed approach can effectively model continuous end-effector actions across large diverse robot datasets, and achieve better generalization performance. 
Extensive experiments demonstrate the effectiveness and generalization of Diffusion Transformer Policy on Maniskill2, Libero, Calvin and SimplerEnv, as well as the real-world Franka arm, achieving consistent better performance on Real-to-Sim benchmark SimplerEnv, real-world Franka Arm and Libero compared to OpenVLA and Octo. 
Specifically, without bells and whistles, the proposed approach achieves state-of-the-art performance with only a single third-view camera stream in the Calvin task ABC$\rightarrow$D, improving the average number of tasks completed in a row of 5 to 3.6, and the pretraining stage significantly facilitates the success sequence length on the Calvin by over 1.2. 
Project Page: \url{https://zhihou7.github.io/dit_policy_vla/}





\end{abstract}

\section{Introduction}

\begin{figure}
    \centering
    \includegraphics[width=.98\linewidth]{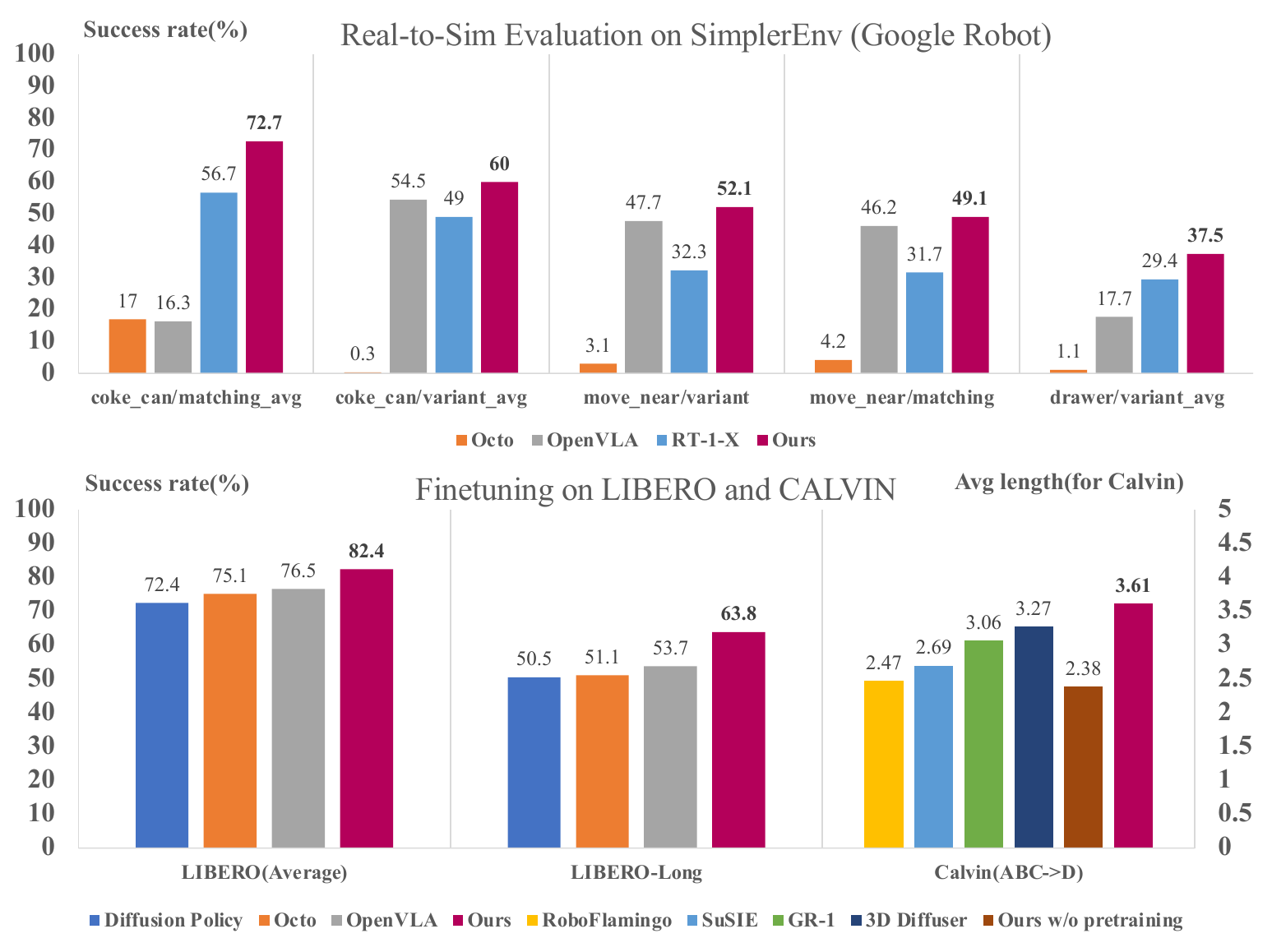}
    \caption{Comparisons between previous state-of-the-art approaches and ours on popular generalist vision-language-action simulation benchmarks. LIBERO~\cite{liu2024libero} and Calvin~\cite{mees2022calvin} are for finetuning generalization, while SimplerEnv~\cite{li24simpler} is to evaluate the generalization to environment variances under Real-to-Sim Benchmark.}
    \label{fig:comp}
\end{figure}

Traditional robot learning paradigm usually relies on large-scale data collected for a specific robot and task, but collecting robot data for generalist tasks is time-consuming and expensive due to the limitations of robot hardware in the real world. 
Nowadays, the foundational models~\cite{openai_chatgpt,openai_gpt4v,openai_dall_e,rombach2021stablediffusion,liu2023llava} in Natural Language Process and Computer Vision, pretrained on broad, diverse, task-agnostic datasets, have demonstrated powerful ability in solving downstream tasks either zero-shot or with a few task-specific samples. 
It is principally possible that a general robot policy exposed to large scale diverse robot datasets improves generalization and performance on downstream tasks~\cite{brohan2022rt,brohan2023rt}. However, it is challenging to train a general robot policy on a large scale of cross-embodiment datasets with diverse sensors, action spaces, tasks, camera views, and environments.

Toward a unified robot policy, existing works directly map visual observation and language instructions to actions with large vision-language-action models for robot navigation~\cite{shah2023gnm,shah2023vint} or manipulation~\cite{brohan2022rt,brohan2023rt,kim2024openvla,team2024octo}, and demonstrate zero-shot or few-shot generalization to new environments. 
Robot Transformers~\cite{brohan2022rt,brohan2023rt,padalkar2023open} present robot policy based on transformer architecture, and demonstrate robust generalization by training on the large scale of Open X-Embodiment Dataset~\cite{padalkar2023open}. Octo~\cite{team2024octo} follows the autoregressive transformer architecture with a diffusion action head, while OpenVLA~\cite{kim2024openvla} discretizes the action space and leverage the pretrained vision-language model to build VLA model exposed to Open X-Embodiment Dataset~\cite{padalkar2023open}. 
Though those Vision-Language-Action (VLA) models~\cite{team2024octo,kim2024openvla} have shown the potential to learn robot policy from the large cross embodiment datasets~\cite{padalkar2023open}, the diversity of robot space among the cross embodiment datasets still limits the generalization.

\begin{figure*}
    \centering
    \includegraphics[width=.83\linewidth]{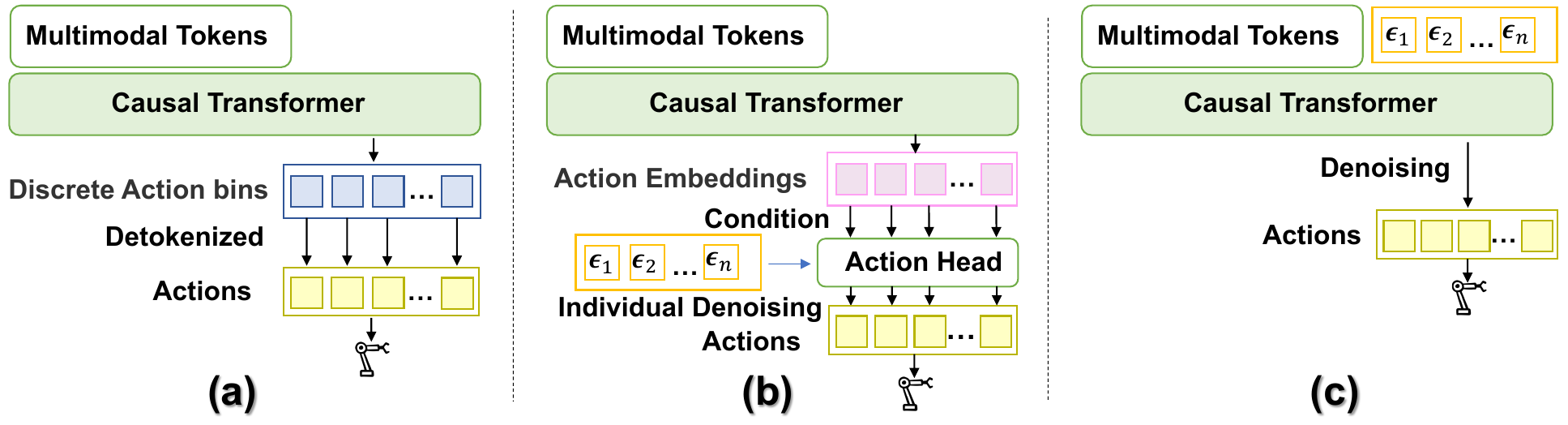}
    \caption{Illustrations of different robot policy architectures. (a) is the common robot transformer architecture with discretization actions, \eg, Robot Transformer~\cite{brohan2022rt,brohan2023rt} and OpenVLA~\cite{kim2024openvla}. (b) is the transformer architecture with diffusion action head which denoises the individual continuous action with a small network condition on each embedding from the causal transformer, \eg, Octo~\cite{team2024octo}. (c) is the proposed Diffusion Transformer architecture that utilizes the large transformer to denoise actions in an in-context conditioning style.}
    \label{fig:enter-label}
\end{figure*}

Recent diffusion policy~\cite{chi2023diffusion,ze20243d,ke20243d,reuss2024multimodal} has shown its stable ability in robot policy learning for single task imitation learning with UNet or cross attention architecture, and diffusion transformer demonstrates its scalability in multi-modal image generation~\cite{peebles2023scalable}. Specifically, Octo~\cite{team2024octo} presents a generalist policy that denoises the action with a small MLP network conditioned on a single embedding of auto-regressive multi-modal transformer. However, the robot space of large-scale cross-embodiment datasets contains various cameras views and diverse action spaces, which poses a significant challenge for a small MLP to separately denoise each continuous action conditioned on a single action head embedding. Meanwhile, previous diffusion policies~\cite{team2024octo,reuss2024multimodal,ke20243d} first fuse historical image observations and instruction into embeddings before the denoising process, which might limit action denoising learning since action anticipation usually relies directly on detailed historical observations rather than fused embedding.


In this paper, we design a Diffusion Transformer architecture for generalist robot policy learning. 
Similar to previous robot transformer models~\cite{brohan2022rt,brohan2023rt,padalkar2023open,team2024octo,kim2024openvla}, we leverage the transformer as our base module to retain the scalability on the large-scale cross-embodiment datasets. 
Different from~\cite{brohan2022rt,brohan2023rt,padalkar2023open,team2024octo,kim2024openvla}, we present an in-context conditional diffusion transformer architecture to denoise the action chunks (\ie, action sequence), rather than utilizing a small shared MLP to separately denoise each action embedding to continuous actions as illustrated in Figure~\ref{fig:enter-label}. Meanwhile, the in-context conditional design enables the action denoising learning directly condition on each image observation patches, which supports the denoising model in perceiving the subtle nuances (\eg, action delta) in historical visual observations. Also, DiT Policy retains the scalability of transformer for diffusion, allowing for more effective generalization from large and diverse embodiment datasets.





In a nutshell, we present a Diffusion Transformer Policy, that incorporates a causal transformer as an in-context conditional diffusion backbone and denoise continuous action chunks. 
Extensive experiments demonstrate DiT Policy achieves considerably better performance on two large-scale Sim datasets, Maniskill2 (novel camera views) and Calvin, compared to baselines. 
Meanwhile, the proposed model trained on the Open X-Embodiment Dataset achieves better generalization performance compared to Octo and OpenVLA on the Real Franka and Libero.

\section{Related Work}

{\bf Diffusion Policy}
Denoise diffusion techniques~\cite{ho2020denoising,rombach2022high,dhariwal2021diffusion,peebles2023scalable,videoworldsimulators2024} are pioneering image generation, and recent Diffusion Policy~\cite{liang2024skilldiffuser,wang2024one,cao2024mamba,wang2024sparse,chen2024diffusion,chi2023diffusion,ze20243d, ke20243d,reuss2024multimodal} has exhibited a powerful ability in modeling multimodal actions compared to previous robot policy strategies. Current diffusion policy approaches usually follow an Unet structure or a shallow cross-attention network for a single manipulation task, leaving large-scale multimodal diffusion policy poorly investigated. For example, 3D diffusion Policy~\cite{ze20243d} denoises policy conditioned on a 3D point cloud, while 3D diffuser actor~\cite{ke20243d} proposes a 3D diffusion strategy based on point cloud with cross-attention. Differently, we present a scalable in-context conditioning diffusion transformer architecture, which directly conditions on each historical observation.  Recent generalist policy Octo~\cite{team2024octo} conditions the denoise process on the embedding from the Transformer model with a small MLP diffuser. By contrast, the diffuser in Diffusion Transformer Policy is a large Transformer architecture.

{\bf Generalist Robot Policies}
Language-conditioned policy~\cite{reuss2023goal,lynch2020language,ha2023scaling,myers2023goal,zhang2022language,chen2023playfusion} is more suitable and general for real applications, and the embodiment community has shown increasing interest in generalist robot policy with foundational multi-modal models for both robot navigation~\cite{shah2023gnm,shah2023vint,yang2024pushing,sridhar2024nomad,huang2023embodied,blessing2024information} and manipulation~\cite{bousmalis2023robocat,brohan2022rt, shah2023mutex, shridhar2023perceiver,brohan2023rt,padalkar2023open,kim2024openvla, team2024octo,driess2023palm, fang2023rh20t,pearce2023imitating,reuss2023goal,lynch2020learning,xian2023unifying,bharadhwaj2024roboagent,xiao2022robotic,mees2022matters,karamcheti2023language,scheikl2024movement}. Recent approaches~\cite{brohan2022rt, brohan2023rt, padalkar2023open, kim2024openvla, team2024octo} aim to achieve generalist policy with scalable Vision-Language-Action models. We follow this paradigm to approach generalist and adaptive robot policy. ~\cite{brohan2022rt, brohan2023rt, padalkar2023open, kim2024openvla} construct the action token by discretizing each dimension of the robot actions separately into 256 bins. However, this discretization strategy incurs internal deviation in robot execution. Unlike those methods, we present a Diffusion Transformer Generalist Policy, which denoises the continuous actions with a large Transformer model. The proposed approach retains the scalability of the Transformer and meanwhile facilitates the modeling of cross-embodiment action chunk representations. Meanwhile, the Diffusion Transformer Policy aligns robot action together with the language instructions and image observations as an in-context conditional style. 






\section{Method}

We describe the architecture of diffusion transformer policy in this section, a DiT-based generalist diffusion policy that can be adapted to new environments and embodiments.

\begin{figure*}[h]
  \centering
  \includegraphics[width=0.90\linewidth]{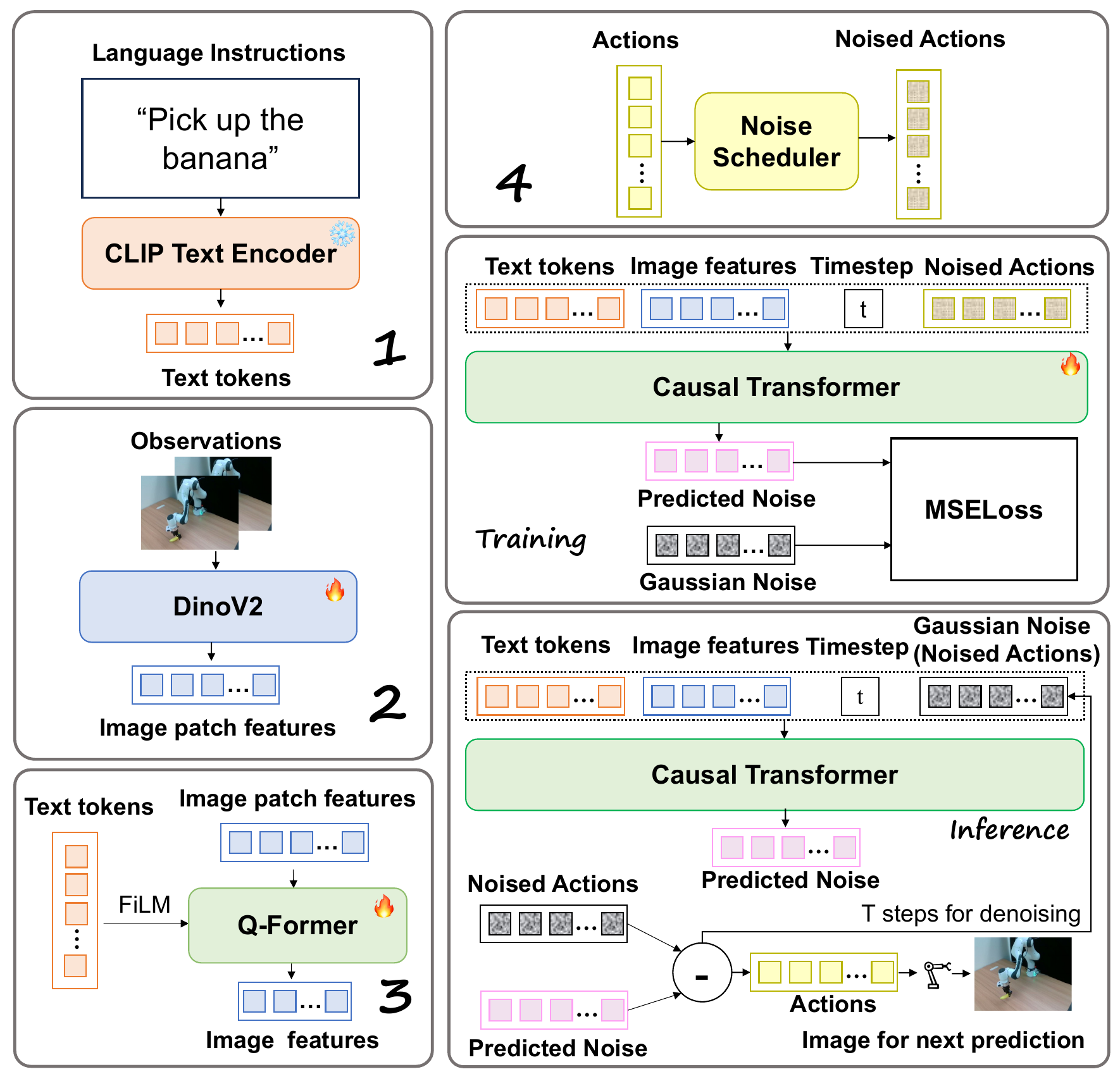}
  \caption{Our model is a Transformer diffusion structure. The model first incorporates a pretrained CLIP network to obtain instruction tokens. Meanwhile, we use the DINO-V2~\cite{oquab2023dinov2} model to encode image observations, followed by a Q-Former to query observation features for each image observation. Next, we concatenate instruction tokens, image observation features, timestep, and noised action together to construct a token sequence as the input for transformer network to denoise the raw actions.}
  \label{fig:main}
\end{figure*}

\subsection{Architecture}

\noindent{\bf Instruction Tokenization}. The language instructions are tokenized by a frozon CLIP~\cite{radford2021learning} model.

\noindent{\bf Image observation Representations}. The image observations first pass into the DINOv2~\cite{oquab2023dinov2} to obtain the image patch features. Note that DINOv2 is trained on the web data which is different from the robot data, we thus jointly optimize the DINOv2 parameters together with Transformers through an end-to-end way.

\noindent{\bf Q-Former}. To reduce the computation cost, a Q-Former~\cite{li2023blip} together with FiLM~\cite{perez2018film} conditioning is incorporated to select image features from the patch features of DINOv2~\cite{oquab2023dinov2} by instruction context.

\noindent{\bf Action Preprocess}. We use the end-effector action and represent each action with a 7D vector, including 3 dimensions for the translation vector, 3 dimensions for the rotation vector, and a dimension for the gripper position. To align the dimension with image and language tokens, we simply pad the continuous action vector with zeros to construct the action representation. We only add the noise into the 7D action vector during denoise diffusion optimization.


\noindent{\bf Architecture}. Our core design is the Diffusion Transformer structure~\cite{peebles2023scalable} which denoises action token chunks, instead of each single action token, conditioned directly on image observation and instruction tokens by an in-context conditioning style with a causal transformer, \ie, we simply concatenate image features, language tokens, and timestep embedding in the front of the sequence, equally treating the noisy action from the instruction tokens as illustrated in Figure~\ref{fig:main}. This design retains the scaling properties of transformer networks, and allows the denoising learning conditioned directly on image patches, thus facilitating the model capture the detailed action changes in historical observations. The model, conditioned on language instructions and image observations with the causal transformer structure, is supervised by the noise that we add to the continuous actions. In other words, we conduct the diffusion objective directly in the action chunk space with a large transformer model, differently from a shared diffusion action head with a few MLP layers~\cite{team2024octo}.

The proposed Diffusion Transformer Policy is a general design that can be scaled to different datasets, and demonstrates excellent performance. Meanwhile, we can also add additional observation tokens and input into the transformer structure. Appendix A provides more details.


\subsection{Training Objective}
In our architecture, the denoising network $\epsilon_{\theta}(\vx^t,c_{obs}, c_{instru}, t)$ is the entire causal transformer, where $c_{obs}$ is the image observation, $c_{instru}$ is the language instruction, and $t\in {1,2,...T}$ is the step index in our experiments. During the training stage, we sample a Gaussian noise vector $\vx^{t}\in \mathcal{N}(\mathbf{0}, \mI)$ at timestep $t$, where $T$ is the number of denoising timesteps, and add it to action $\va$ as $\hat{\va}$ to construct the noised action token, finally predicting the noise vector $\hat{\vx}$ based on the denoising network $\epsilon_{\theta}(\hat{\va},c_{obs}, c_{instru}, t)$, where $t$ is randomly sampled during training. We optimize the network with MSE loss between $\vx^t$ and $\hat{\vx^t}$.

To generate an action, we apply $T$ steps of denoising with the optimized transformer architecture $\epsilon_{\theta}$ from a sampled gaussian noise vector $\vx^{T}$ as follows,

\begin{equation}
\vx^{t-1} = \alpha(\vx^{t} - \gamma\epsilon_{\theta}(\vx^t,c_{obs}, c_{instru}, t) + \mathcal{N}(\mathbf{0}, \sigma^2\mI)).
\end{equation}

%

where $\alpha$, $\gamma$, $\sigma$ is the noise scheduler~\cite{ho2020denoising}. In our experiments, $\epsilon_{\theta}$ is to predict the noise that adds to the action.

\subsection{Pretraining Data}
To evaluate the proposed Diffusion Transformer Policy, we choose Open X- Embodiment datasets~\cite{padalkar2023open} for pretraining the model. We mainly follow~\cite{team2024octo,kim2024openvla} to choose the datasets and set the weights for each dataset. We normalize the actions similar to~\cite{padalkar2023open} and filter out outlier actions in the dataset. Additional details are provided in Appendix B.



\subsection{Pretraining Details}
 We devise the proposed Diffusion Transformer architecture and evaluate the pretraining approach in the large cross-embodiment datasets~\cite{padalkar2023open}. We use the DDPM~\cite{ho2020denoising} diffusion objective in the pretraining stage with $T=1000$ for the Open X-Embodiment dataset~\cite{padalkar2023open}, while we set $T=100$ with DDIM~\cite{song2020denoising} for zero-shot evaluation to accelerate the inference. According to the preliminary experiment from Maniskill2~\cite{gu2023maniskill2}, we use 2 observation images and predict 32 action chunks. We train the network with AdamW~\cite{loshchilov2017decoupled} by 100,000 steps. We set the learning rate of the casual transformer and Q-Former as 0.0001, the learning rate of DINOv2 as 0.00001, and the batch size as 8902. More pretraining details are provided in the Appendix A.

\begin{table}
\setlength\tabcolsep{1.5pt}

\caption{Comparison with RT-1-X~\cite{brohan2022rt}, Octo-base~\cite{team2024octo} and OpenVLA-7B~\cite{kim2024openvla} on SimplerEnv (average variance and matching results of Google Robot~\cite{brohan2022rt}). The results are reported as success rate.}
\label{tab:simplerenv}
\small
\begin{center}
\begin{tabular}{l|cc|cc|cc}
\toprule
\multirow{2}{*}{\bf Method}  &\multicolumn{2}{c}{\bf coke$\_$can} & \multicolumn{2}{c}{\bf move$\_$near } & \multicolumn{2}{c}{\bf drawer}  \\
 & { match } & {variant} & {match} & {variant} & {match} & {variant} \\
\hline 

RT-1-X~\cite{brohan2022rt} & 56.7\% & 49.0\% & 31.7\% & 32.3\% & \bf 59.7\% & 29.4\%  \\
Octo-Base~\cite{team2024octo} & 17.0\% & 0.6\% & 4.2\% & 3.1\% &  22.7\% & 1.1\%  \\
OpenVLA-7B~\cite{kim2024openvla} & 16.3\% & 54.5\% & 46.2\% & 47.7\% & 35.6\% & 17.7\% \\
DiT Policy(Ours) & \bf 72.7\% & {\bf 60.0\%} & \bf 56.7\% & \bf 57.5\% & 46.3\% & \bf 37.5\% \\

\bottomrule
\end{tabular}

\end{center}
\end{table}

\section{Experiments}

We evaluate the proposed methods with two baselines in three environments. We leverage Maniskill2 to present the ability of Diffusion Transformer Policy on large scale novel view generalization. Meanwhile, we demonstrate the generalization of the pretrained Diffusion Transformer Policy on CALVIN benchmark. Lastly, we further show the generalization of DiT Policy on Real Franka Arm.

\subsection{Baselines}

\noindent{\bf Discretization Action Head} We implement the RT-1~\cite{brohan2022rt} style baseline models with a similar structure as ours. We keep the Instruction Tokenization and Image backbone. Different from ours, we discretize each dimension of the action into 256 bins~\cite{brohan2022rt}, and leverage the transformer network to predict the action bin indexes. Following~\cite{brohan2022rt,brohan2023rt}, we use cross-entropy loss to optimize the network.

\noindent{\bf Diffusion Action Head} We also implement a diffusion action head strategy ~\cite{team2024octo}. Specifically, we utilize a three-layer MLP network as our denoising network condition on the output of each action token embedding by the same transformer architecture as ours. Notably, this baseline has more parameters (the additional MLP) compared to DiT policy.


\subsection{SimplerEnv}

SimplerEnv~\cite{li24simpler} is a Real-to-Sim platform for evaluating the policy learned from real robot data with a simulation platform. In this section, we compare with the popular generalist policy trained on Open X-embodiment dataset, including RT-1-X~\cite{brohan2022rt,padalkar2023open}, Octo~\cite{team2024octo} and OpenVLA~\cite{kim2024openvla}, on Google Robot Simulation with different variances. We follow the evaluation from SimplerEnv~\cite{li24simpler} for a fair comparison, which includes ``pick up coke can'', ``move an object near to others'', ``open drawer'', ``close drawer''.

Table~\ref{tab:simplerenv} demonstrates that the proposed approach achieves a strong generalization performance under different variances, including background, texture, objects, spatial positions, and so on. The variance experiments show the robust of Diffusion Transformer Policy.

\subsection{CALVIN}

CALVIN (Composing Actions from Language and Vision)~\cite{mees2022calvin} is an open-source simulated benchmark to learn long-horizon language-conditioned tasks. CALVIN~\cite{mees2022calvin} includes four different scenes tagged as ABCD and presents a novel scene evaluation benchmark, ABC$\rightarrow
$D, \ie, trained on environments A, B, and C and evaluated on environment D. The goal of CALVIN is to solve up to 1000 unique sequence chains with 5 distinct subtasks. The benchmark requires successfully solving the task sequence with 5 continuous subtasks, and one of the important evaluation indicators is the success sequence length.

\textit{Setup}. In this section, we utilize CALVIN (ABC$\rightarrow$D) to evaluate the novel task generalization of Diffusion Transformer Policy architecture. Specifically, we directly apply the proposed method on CALVIN with a single static RGB camera and predict the end-effector action, including 3 dimensions for translation, 3 dimensions for  Euler angles rotation and 1 dimension for gripper position (open or close). We evaluate Diffusion Transformer Policy and Diffusion Action Head~\cite{team2024octo} on CALVIN, and leverage the pretrained model on Open X-Embodiment to initialize the model for CALVIN.

\textit{Optimization Details}.  While training Calvin, 2 history images are used as input. For each iteration, the model predicts 10 future frames supervised by MSE loss. An AdamW optimizer is used together with a decayed learning rate with half-cycle cosine scheduler after several steps of warming up. The learning rate is initialized as 1e-4. We use 4 NVIDIA A100 GPUs(80GB) to train the model for 15 epochs with a global batch size of 128.

\textit{Comparisons}. Table~\ref{tab:calvin} presents the comparisons with previous methods on Calvin and the proposed methods. Without whistles and bells, the proposed Diffusion Tranformer Policy achieves the state-of-the-art results. Particularly, we only use RGB camera stream for observation. The superior demonstrates the effectiveness of Diffusion Transformer Policy. Meanwhile, the pretraining on Open X-Embodiment Datasets significantly facilitates the performance by 1.23, which demonstrates the transferability of Diffusion Transformer Policy. By contrast, the performance of diffusion action head is worse than Diffusion Transformer Policy by 0.45, though we load similar pretraining weights for diffusion head architecture.
DiT Policy can perceive visual subtle nuances for long-horizon tasks, and scale across different environments, \eg, transferring the knowledge from the diverse real datasets to the CALVIN dataset.

 \begin{table*}[t]
 \centering
 \caption{The comparisons with state-of-the-art approaches on Calvin Benchmark under success rate and average success length. MDT~\cite{reuss2024multimodal} is from issue 9 of its GitHub repo and GR-MG~\cite{li2025gr}. `S-RGB' indicates Static RGB, `G-RGB' indicate Gripper RGB. `S-RGBD' indicate Static RGB-D, `G-RGBD' indicates Gripper RGB-D, `P' is the observation arm position `Proprio', `Cam' indicates camera parameters. `Pretraining' indicates pretraining dataset or multi-modal model.}
  \label{tab:calvin}
  \small
  \begin{tabular}{c|c|ccccc|c}
\toprule    
    \multirow{2}{*}{Method} & \multirow{2}{*}{Input} & \multicolumn{6}{|c}{No. Instructions in a Row (1000 chains)} \\ 
    \cline{3-8}
    & & 1 & 2 & 3 & 4 & 5 & Avg.Len. \\ \hline
    MDT*~\cite{reuss2024multimodal} & & 61.7\% & 40.6\%& 23.8\%& 14.7\%& 8.7\% &  1.54 \\
    SPIL ~\cite{zhou2024language} &	S-RGB,G-RGB & 74.2\% &	46.3\%& 	27.6\%&	14.7\%&	8.0\%&	1.71 \\
     RoboFlamingo~\cite{li2023vision}  &S-RGB,G-RGB &	82.4\%	&61.9\%	& 46.6\%&	33.1\%	&23.5\%	& 2.47 \\
SuSIE~\cite{black2023zero}&	 S-RGB &	87.0\%&	69.0\%	& 49.0\%	&38.0\%	 &26.0\%& 	2.69 \\
GR-1~\cite{wu2023unleashing} & S-RGB,G-RGB,P &	85.4\%	&71.2\%	 & 59.6\%	&49.7\%	& 40.1\%&	3.06 \\
3D Diffuser~\cite{ke20243d} &	S-RGBD,G-RGBD,P,Cam&	92.2\%	&78.7\%	& 63.9\%	& 51.2\%	& 41.2\%	& 3.27 \\
\hline

diffusion head w/o pretrain  & S-RGB &  75.5\%  & 44.8\%  & 25.0\% & 15.0\% & 7.5\% & 1.68 \\
diffusion head  & S-RGB &  94.3\%  & 77.5\%  & 62.0\% & 48.3\% & 34.0\% & 3.16 \\
Ours w/o pretrain & S-RGB &  89.5\%  & 63.3\%  & 39.8\% & 27.3\% & 18.5\% & 2.38 \\
    Ours & S-RGB &  {\bf 94.5\%}  & {\bf 82.5\%}  & {\bf 72.8\%} & {\bf 61.3\% } & {\bf 50.0\%} & {\bf 3.61} \\

%


\bottomrule
\end{tabular}

\end{table*}

\begin{table*}[t]
\caption{Comparison with Diffusion Policy\cite{chi2023diffusion}, Octo\cite{team2024octo} and OpenVLA\cite{kim2024openvla} on LIBERO\cite{liu2024libero}. Besides the results of our model, all of the other results are from \cite{kim2024openvla}. The results are a little different with we reported in the paper, this is because that we find a more suitable scheduler for finetuning and make a little increasing on the performance.}
\label{tab:libero}
\small
\begin{center}
\begin{tabular}{l|cccc|c}
\toprule
{\bf Method}  & {\bf LIBERO-SPATIAL} & {\bf LIBERO-OBJECT} & {\bf LIBERO-GOAL} & {\bf LIBERO-LONG} & {\bf Average} \\
\hline 
Diffusion Policy from scratch\cite{chi2023diffusion} & 78.3\% & 92.5\% & 68.3\% & 50.5\% & 72.4\% \\
Octo fine-tuned\cite{team2024octo} & 78.9\% & 85.7\% & 84.6\% & 51.1\% & 75.1\% \\
OpenVLA fine-tuned\cite{kim2024openvla} & \bf 84.9\% & 88.4\% & 79.2\% & 53.7\% & 76.5\% \\
DiT Policy fine-tuned(Ours) & 84.2\% & \bf 96.3\% & \bf 85.4\% & \bf 63.8\% & \bf 82.4\% \\

\bottomrule
\end{tabular}
\end{center}

\end{table*}

\subsection{LIBERO}
\label{sec:analysis:libero}

LIBERO is a general benchmark focused on knowledge transfer in multitask and lifelong robot learning problems\cite{liu2024libero}. This benchmark consists of four sub-datasets, which are called LIBERO-SPATIAL, LIBERO-OBJECT, LIBERO-GOAL and LIBERO-100,  respectively. These sub-datasets are used to test different abilities of the models. LIBERO-SPATIAL tests the ability of spatial relationship understanding, containing data of the same object set but different layouts. LIBERO-OBJECT tests the ability of object transferring, containing data of the same layouts but different object sets. LIBERO-GOAL tests the ability of task understanding and transferring, containing data of the same object sets and layouts but different tasks. LIBERO-100 can further divide into LIBERO-90 and LIBERO-10(Also called LIBERO-LONG), which are used for policy pretraining and long-horizon tasks testing, respectively. LIBERO-100 consists of data of diverse objects, layouts and backgrounds. In the latest version of OpenVLA\cite{kim2024openvla}, they compares the performance of several models on LIBERO. We follow OpenVLA and make a comparison with these models. We first pretrain our model on the Open X-Embodiment\cite{padalkar2023open} then full-finetune it on LIBERO. We use the dataset that slightly modified by OpenVLA\cite{kim2024openvla} in order to show a fair comparison. The results are shown in Table~\ref{tab:libero}. Besides the results of our model, all of the other results are from \cite{kim2024openvla}. We achieve a better performance on most of the sub-datasets of LIBERO and make a nearly 6\% increasing on the average success rate. Specially, we make great progress on the LIBERO-LONG, which means a great potential for completing the long-horizon tasks of our model.

\subsection{Maniskill2}

{\bf Maniskill2}~\cite{gu2023maniskill2} is the next generation of the SAPIEN Maniskill benchmark~\cite{mu2021maniskill}, which is widely used to evaluate the generalized manipulation ability of the embodied models. It contains 20 different manipulation tasks families and over 4 million demonstration frames with different settings, including rigid/soft body, single/dual arm, etc. Maniskill2 also provides a fast and easy way to change the camera view and replay the trajectories. It is useful for the researchers working on generalized policy. 

\textit{Setup}. In our experiments, we select 5 tasks (PickCube-v0, StackCube-v0, PickSingleYCB-v0, PickClutterYCB-v0, PickSingleEGAD-v0) from Maniskill2, and then construct a camera pool with 300,000 random cameras, then sample 20 cameras from the camera pool to render a trajectory each time, and finally obtain about 40K trajectories totally. Given that the number of trajectories is large, we thus train the network on Maniskill2 from scratch. Moreover, we split the dataset into training set and validation set according to a ratio of 19:1. During the splitting, it is guaranteed that the single trajectory rendered under different camera views will appear in either training set or validation set in order to avoid data leaking. Specially, there are 74 different categories to pick and place in the task family PickSingleYCB-v0. In addition to the mentioned rules, we ensure each category can be found in both training set and validation set. After that, we sample 100 trajectories for each task family randomly from the validation set, constructing a close loop evaluation dataset with 500 trajectories in total. While training, considering the balance between different task families, we adjust the number of data pieces to the same by simply copying the trajectories from task family with fewer trajectories originally. The ability of the model is measured by the success rate of each task family while executing the close loop evaluation dataset.

\textit{Optimization Details}. We optimize the network with AdamW~\cite{loshchilov2017decoupled} by 50,000 steps on Maniskill2 and we set the learning rate as 0.0001. The number of training timesteps $T$ is 100 in Maniskill2 and the global batch size is 1024.

\textit{Comparisons}. Table~\ref{tab:maniskill_main} compares the proposed method with discretized action head~\cite{brohan2022rt, brohan2023rt} and diffusion action head~\cite{team2024octo}. The experiments demonstrate Diffusion Transformer Policy achieves better results compared to Discretization Action Head strategy~\cite{brohan2022rt} under the large scale novel view maniskill2 benchmark. Meanwhile, Diffusion Transformer Policy demonstrates better performance in more complex tasks, \eg, Diffusion Transformer Policy improves diffusion action head~\cite{team2024octo} by 20\% in task PickSingleYCB and by 12\% task PickClutterYCB. Those experiments show that Diffusion Transformer Policy achieves better scalability in the large scale diverse datasets, and meanwhile achieves better generation in camera view generalization. 

\begin{table*}[t]
\caption{Comparision with two baseline methods on Maniskill2 under success rate. SingleYCB indicates PickSingleYCB, ClutterYCB indicates PickClutterYCB, SingleEGAD indicates PickSingleEGAD. Disc ActionHead indicates Discretized Action Head strategy~\cite{brohan2022rt}, while Diff ActionHead shows Diffusion Action Head~\cite{team2024octo}.}
\label{tab:maniskill_main}
\small
\begin{center}
\begin{tabular}{l|c|ccccc}
\toprule
{\bf Method}  & {\bf All} & {\bf PickCube} & {\bf StackCube} & {\bf SingleYCB} & {\bf ClutterYCB} & {\bf SingleEGAD} \\
\hline 
Disc ActionHead & 30.2\% & 41.0\% & 33.0\% & 22.0\% & 1.0\% & 54.0\% \\
Diff ActionHead & 58.6\% &  {\bf 86.0\%} &  76.0\% &  37.0\% &  24.0\% &  70.0\% \\
DiT Policy(ours) & {\bf 65.8\%} &  79.0\% &  {\bf 80.0\%} &  {\bf 62.0\%} &  {\bf 36.0\%} &  {\bf 72.0\%}  \\

\bottomrule
\end{tabular}
\end{center}

\end{table*}




\subsection{Real Franka Arm}
We finally evaluate the proposed method by pretraining on Open X-Embodiment Datasets~\cite{padalkar2023open}, and evaluate it in our Franka Arm environment under zero-shot generalization, 10-shot generalization, and in domain finetuning generalization. We meanwhile compare it by finetuning on Libero~\cite{liu2024libero}.

\begin{figure}
    \centering
    \includegraphics[width=0.95\linewidth]{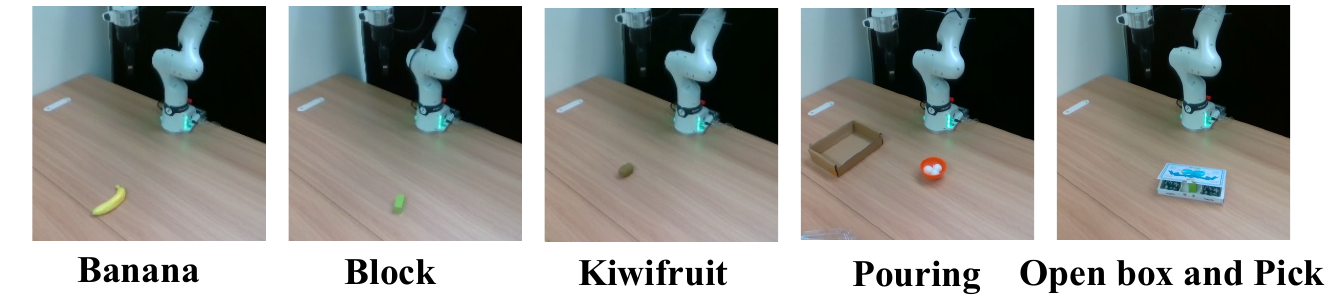}
    \caption{Illustration of Franka environment and exampled tasks.}
    \label{fig:tasks}
\end{figure}

\textit{Setup}. We set up the franka on the table with a black background. Meanwhile, we use a single third-person RGB camera about 1.5 meters away from the Franka Arm. Please refer to Figure~\ref{fig:tasks} for the visualized demonstration. Considering that the environment of our Franka setup is different from the scenes in the pretraining data Open X-Embodiment~\cite{stone2023open}, we mainly evaluate the proposed method on out-of-the-box generation and few-shot generation. In our experiments, we evaluate each model with the same scene, and the object is placed in 9 similar positions in a 9-grid format in front of the franka arm. Meanwhile, we maintain a small variance in those positions placing the objects for evaluation.

To evaluate the ability of quick finetuning with a few training samples, we set two pick and place tasks, and further collect five complex manipulation tasks, \ie, `Pick up the bowl and pour the balls into the box', `Open the box and pick up the block within the box', `pick up the banana and move it into the pen container', `stack the bowls', and `pick up the cup and pouring the coffee beans from the cup into the bowl'. We collect 10 samples for each task for \textit{ 10-shot finetuning generalization}. Figure~\ref{fig:tasks} presents the scenes and tasks. The image in Figure~\ref{fig:tasks} is the model input. Our real-world environment is challenging since the object (less than 20mm) is small compared to the whole scene. We also present a simple evaluation for in-domain evaluation for comparing different action strategies under in-domain generalization with 5 pick and place tasks, including `green block', `kiwifruit', `banana', `the tiny green block', `pink block'. Meanwhile, we collect 50 trajectories for each task in the first three tasks, while leaving the remaining two tasks (`Pick up the tiny green block' and `Pick up the pink block') for out-of-distribution evaluation.

\textit{Finetuning details.} In our experiments, we finetune the proposed method on the real Franka Arm with Lora~\cite{hu2021lora} and AdamW~\cite{loshchilov2017decoupled} for 10,000 steps. We set the number of timesteps as 100 for DDPM~\cite{ho2020denoising}, and batch size as 512.

{\bf 10-shot Finetuning Generalization} We directly take the models pretrained on Open X-Embodiment to evaluate 10-shot generalization in our environments. Here, to maintain the consistency with OpenVLA~\cite{kim2024openvla}, we finetune the network with one observation and one step prediction. We compare the proposed method with OpenVLA~\cite{kim2024openvla} and Octo-base~\cite{team2024octo} models. Besides, we evaluate it on long horizon tasks. Table~\ref{tab:franka_10_shot} presents the proposed Dit Policy archives consistent improvement compared to Octo and OpenVLA. Specifically, the proposed DiT policy demonstrates slight improvement on Pickup tasks, while the proposed DiT policy can achieve clear better performance on all more complex tasks. For those long-horizon tasks, we can find OpenVLA can
complete the first task effectively while failing to under-
stand the long-horizon task. For example, it completely
misunderstands the insert operation, which we illustrate in
the demo video. Differently, Octo is better at completing
tasks with rotations and approaching the second tasks as il-
lustrated with the Demo video. By contrast, the proposed DiT
policy is more robust, and achieves clearly better perfor-
mance in long-horizon tasks. We provide a visualized analysis in Appendix C.1.

\begin{table*}[t]
\caption{Comparision with Octo and Openvla on Real Franka Arm under 10-shot finetuning. Results are reported as success rate. 10-shot finetuning indicates we finetune the model with only 10 samples for each task. `Pourballs' indicates pick up the bowl and pour the balls into the box. `Open \& Pick' indicates `open the box (Step-1) and pick up the block(Step-2) from the box'. `Pick\&Insert' indicates `pick up the banana (Step-1) and insert into the small pen container (Step-2)'. `Stacking Bowls' is `stack the three bowls', in which `Step-1' is the first tacking and `Step-2' is the second tacking. `Pick\& Pouring' is picking the mark cup (Step-1) and pouring the coffee beans from the cup into the bowl (Step-2).}
\label{tab:franka_10_shot}
\small
\begin{center}
\begin{tabular}{l|c|c|c|c|c|c|c|c|c|c|c}
\toprule

\multirow{2}{*}{Method} & \multirow{2}{*}{PickBlock} & \multirow{2}{*}{PickBanana} & \multirow{2}{*}{Pourballs} & \multicolumn{2}{c|}{Open \& Pick } & \multicolumn{2}{c|}{Stacking Bowls} & \multicolumn{2}{c|}{Pick \& Insert } & \multicolumn{2}{c}{Pick \& Pouring} \\
\cline{5-12}
 &  &  &   & {Step-1} & {Step-2} & {Step-1} & {Step-2} &  {Step-1} & {Step-2}  & {Step-1} & {Step-2}\\
\hline
Octo~\cite{team2024octo}  & 7.4\% & 22.2\%  & 14.8\% & 22.2\% & 0 & 10\% & 0 & 20\% & 0 & 10\% & 0\\
OpenVLA~\cite{kim2024openvla}  & 7.4\% & 33.3\% & 18.5\% & 7.4\% & 0  & 40\% & 0 & 80\% & 0 & 20\% & 0 \\
Ours & {\bf 14.8\%} & {\bf33.3}\%  & {\bf 29.6\%} & {\bf 29.6\%} & {\bf 7.4\%}  & {\bf 50\%} & {\bf 12\%} & {\bf 90\%} & {\bf 10\%} & 40\% & 10\% \\



\bottomrule
\end{tabular}

\end{center}
\end{table*}

{\bf In domain Finetuning Generalization}  Table~\ref{tab:franka_few_shot} presents the performance of the proposed DiT Policy compared to baseline methods. We observe different objects demonstrate various performances according to their attributes. The banana is the easiest object to pick up because the banana is longer, while kiwifruit is fat compared to other objects and all models achieve poor performance. The proposed DiT Policy effectively improves the diffusion action head according to Table~\ref{tab:franka_few_shot}. We find the discretized action head baseline achieves poor performance. Meanwhile, the DiT Policy is still able to pick up the novel object (\eg, the pink object) with a low success rate, while the baseline methods totally fail.

\begin{table}[t]
\caption{Comparision on Real-Franka tasks with few-shot fine-tuning about different action strategies. Discretized indicates discretized action head and Diffusion indicates Diffusion action head. We represent the values in success rate. Block is `Pick up the green block', Banana is `Pick up the banana', Kiwi is `Pick up the Kiwifruit', Tiny is unseen object task `Pick up the green tiny block', Oval is unseen object task `Pick up the pink oval block'.}
\setlength\tabcolsep{1pt}
\label{tab:franka_few_shot}
\small
\begin{center}
\begin{tabular}{l|c|ccc|cc}
\toprule
{\bf Method}  & All & { Block } & { Banana } & { Kiwi } & { Tiny } & { Oval } \\
\hline

Discretized~\cite{brohan2022rt} &  19.3\% & 29.6\% & 51.9\% & 14.8\% & 0 & 0 \\
Diffusion ~\cite{team2024octo} & 34.8 \% &  40.7\%  &  85.2\% & 25.9\% & 22.2\% & 0\\

DiT Policy (ours)     & {\bf 46.9\%}  & {\bf 55.6\%} & {\bf 90.3\%} & {\bf 44.4\%} & {\bf 37.0\% } & {\bf 7.4\% }\\

\bottomrule
\end{tabular}
\end{center}

\end{table}

\begin{table}[t]
\setlength\tabcolsep{1pt}
\caption{Ablation on Maniskill2 about the number of history observation images and the length of the trajectory. \#obs indicates the number of history observation images. \#traj shows the length of trajectory, \ie, the sum length of observation and action prediction chunks. PickC indicates PickCube, StackC indicates StackCube, SingleYCB indicates PickSingleYCB, ClutterYCB indicates PickClutterYCB, EGAD indicates PickSingleEGAD.}
\label{tab:maniskill_ab}
\small
\centering
\begin{tabular}{c|c|c|ccccc}
\toprule
{ \#obs } & { \#traj}  & All & { PickC} & { StackC} & { SingleYCB} & { ClutterYCB} & { EGAD}\\
\hline
2 & 2 & 40.8\% &  68.0\% &  54.0\% &  33.0\% &  9.0\% &  40.0\%  \\
2 & 4 & 51.6\% &  81.0\% &  69.0\% &  44.0\% &  11.0\% &  53.0\%  \\
2 & 8 & 62.4\% &  {\bf 89.0} \% &  78.0\% &  54.0\% &  25.0\% &  66.0\%  \\
2 & 16 & 65.6\% &  83.0\% &  {\bf 80.0} \% &  {\bf 70.0} \% &  25.0\% &  70.0\% \\
2 & 32 & {\bf 65.8} \% &  79.0\% &  {\bf 80.0} \% &  62.0\% &  {\bf 36.0} \% &  {\bf 72.0}\% \\
\hline
1 & 32 & 61.6\% &  78.0\% &  76.0\% &  64.0\% &  24.0\% &  66.0\%  \\
1 & 1  & 51.0\% &  79.0\% &  66.0\% &  42.0\% &  19.0\% &  49.0\%  \\
3 & 3 & 35.4\%  &  54.0\% &  49.0\% &  27.0\% &  5.0\% &  42.0\%  \\

\bottomrule
\end{tabular}
\end{table}

\begin{figure}
    \centering
    \includegraphics[width=1.\linewidth]{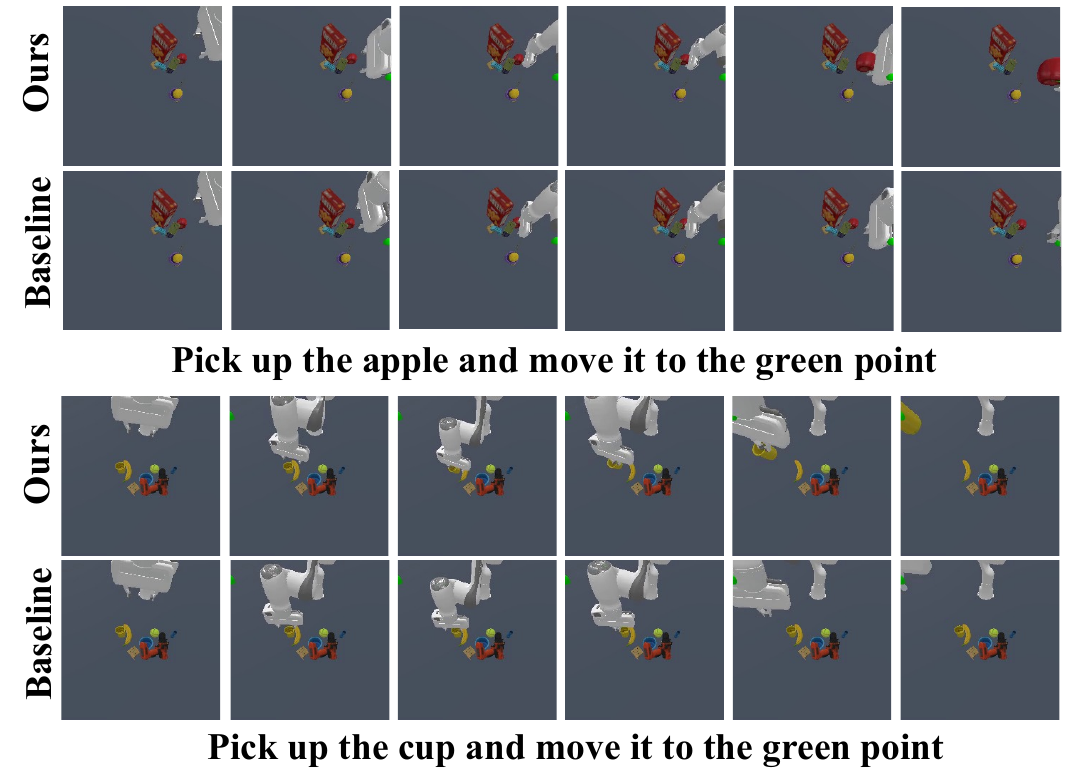}
    \caption{Visualized comparison between Diffusion Transformer Policy and Diffusion Action Head baseline on Maniskill2 (PickClutterYCB). The first raw is Diffusion Tranformer Policy, while the second raw is the baseline method with Diffusion Action Head. }
    \label{fig:vis_comp}
\end{figure}

\subsection{Ablation Study}

In this section, we ablate some of the important designs of the model architecture, including the length of horizon, the length of observation, execution steps for evaluation on Maniskill2. 

{\bf Trajectory length.} The length of action chunks has an important effect on the performance of different tasks. Table~\ref{tab:maniskill_ab} shows that performance increases with increasing trajectory length. Meanwhile, we notice the performance of more complex tasks, \eg, PickClutterYCB, increases significantly with increasing trajectory length, while the easy task, \eg, PickCube, maintains high performance after the trajectory length is greater than 4. Meanwhile, the long horizon optimization significantly facilitates the performance since long horizon optimization is able to provide the target object position and help the model understand the localization of the object. For example, task PickClutterYCB with multiple YCB objects, requires the model to understand which one is the corresponding object.

{\bf Observation length.} In our experiments, we find the length of history observation images also significantly affects the performance. At first, the performance significantly drops when we increase the length of observation history to 3. It might because it is more difficult for the model to converge with more observations since the number of corresponding image tokens also increases. Secondly, we observe using two image observations is more helpful for the performance when the prediction horizon is long. For example, when the length of trajectory is 32, the experiment with two observations achieves better performance. We think two observations can provide the visualized difference between two positions, and the difference of continunous gripper position indicates the action. The visualized difference is beneficial for future action prediction. However, for short horizon, the model majorly learns the projection from current observation to the corresponding actions.


{\bf Execution steps.}
Since the proposed model is able to predict multiple future actions, we can execute multiple steps in one inference. Here, we ablate the effect of execution steps under a model with trajectory length 32 in Table~\ref{tab:maniskill_ab1}. The ablation study shows that the short execution steps are slightly better longer execution steps, \ie, the farther away from the current frame, the worse the prediction quality.


\begin{table}[t]
\caption{The effect of the number of execution steps on Maniskill2. \#steps indicates the number of steps that we execute each prediction.}
\label{tab:maniskill_ab1}
\small
\centering
\begin{tabular}{l|c|c|c|c|c}
\toprule
{ \#steps } & 1 &  2  & 4  & 8 & 16 \\
\hline
All & {\bf 61.6}\% & 60.8 \% & 60.6 \% & 60.0 \% & 58.0 \%  \\
\bottomrule
\end{tabular}
\end{table}

\subsection{Visualized Comparison}

\noindent{\bf Maniskill2}. The proposed Diffusion Transformer Policy is able to model better action sequences. We conduct visualized analysis between the proposed method and the diffusion action head baseline on Maniskill2 in Figure~\ref{fig:vis_comp}. We select two trajectories from PickClutterYCB task, which is the most challenging task in Maniskill2. Figure~\ref{fig:vis_comp} presents the grasp position is significantly important for picking up successfully, and the main reason that the baseline fails to pick up is the wrong grasp position. Meanwhile, we observe the major challenge of task PickClutterYCB is the grasping position prediction, especially when the target object is near by other objects. Compared to the diffusion action head baseline, Diffusion Transformer Policy is able to predict better action chunks for correctly picking the object with a suitable end-effector pose.

\noindent{\bf Real Franka Arm}. We also illustrate the comparison between the Diffusion Transformer Policy and diffusion action head baseline on real Franka Arm in Figure~\ref{fig:vis_comp_real}. We demonstrate the experimental results under few-shot finetuning setting. We find the proposed method achieves better action prediction when the Gripper is approaching the object and finally picks up the small green object successfully, while the baseline fails to pick up due to the inaccurate grasp position. Meanwhile, we observe that failures are usually caused by a tiny position bias and we can not even directly discriminate the position by eyes from the image. For those cases, we argue the diffusion transformer policy has learnt better grasp position during the pretraining stage, and thus reduce failure rate due to the wrong grasp pose, while it is difficult for the diffusion action head. We demonstrates more comparisons to analyze the challenges in the Franka Arm.

\begin{figure}
    \centering
    \includegraphics[width=1.\linewidth]{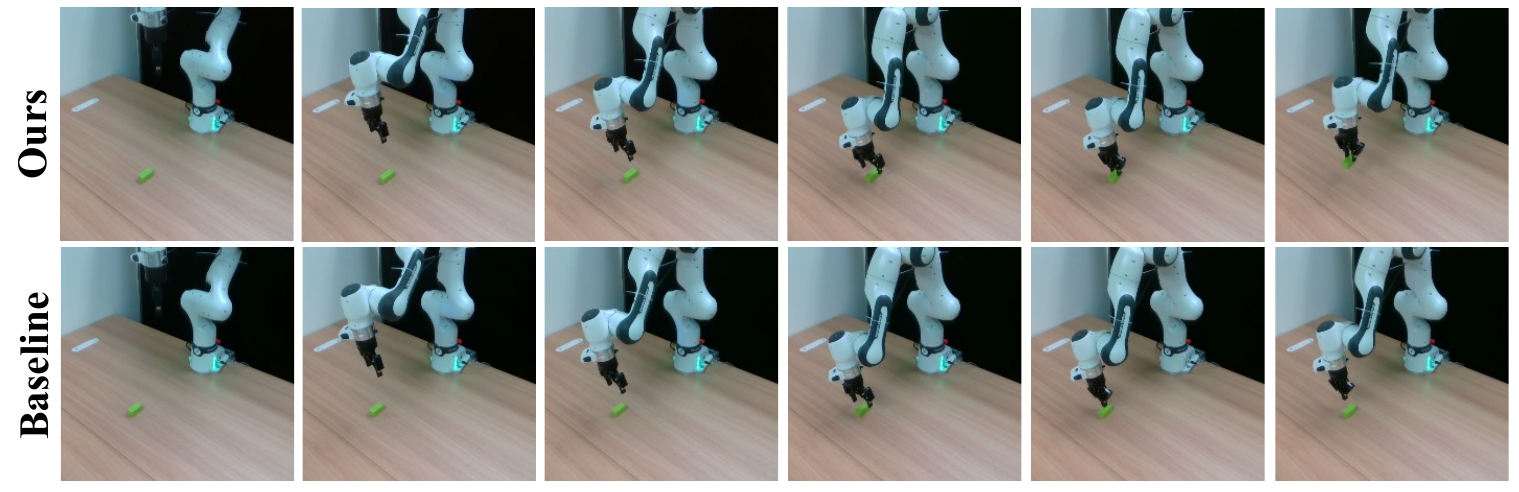}
    \caption{Visualized comparison between Diffusion Transformer Policy and Diffusion Action Head strategy on Real Franka Arm (Pick up the green block). The first raw is DiT Policy, while the second raw is the baseline method with Diffusion Action Head. }
    \label{fig:vis_comp_real}
\end{figure}

\section{Conclusion and Discussion}

In this paper, we present a Diffusion Transformer architecture for generalist robot learning, named as Diffusion Transformer Policy. Diffusion Transformer Policy directly utilizes the large transformers as a denoising network to denoise the continuous actions conditioned on language instruction and image observations. The proposed architecture retains the scale attribute of the transformer, thus is capable of generalizing to different datasets with a unified architecture. Extensive experiments on SimplerEnv, Libero, Maniskill2, CALVIN, real Frank Arm demonstrate the effectiveness of the proposed method. Particularly, the proposed approach achieves state-of-the-art performance in CALVIN (ABC$\rightarrow$D) with only a single observation, and achieves consistent improvement compared to OpenVLA and Octo on SimplerEnv.

A limitation of the Diffusion Transformer Policy is that it requires multiple denoising steps during inference, which will impede the inference speed in the real application. In this paper, we focus on the modeling of complex and diverse actions. We think it is possible to improve the finetuning strategy with a few denoising steps to accelerate the inference speed. 

{
    \small
    \bibliographystyle{ieeenat_fullname}
    \bibliography{main}
}

\appendix
\clearpage
\setcounter{page}{1}
\maketitlesupplementary

\section{Additional Training and Model Details}
\label{sec:details}

The structure of our Model is shown in Figure 2. In this section, we will talk about some extra details  of our model. The language instruction is encoded by a pretrained clip model and freeze the encoder in the training loop. We then resize the input images into 224$*$224 and feed it into a pretrained ViT model. The selected ViT is the base version of DinoV2. All the parameters in DinoV2 are trained. After the above process, we use a Q-Former to decrease the size of image features. The Q-Former is from scratched with a depth of 4. In each block, we insert the text token as a FiLM Condition to get the image features containing language information. The query length of the image features will reduce to 32 when out of the Q-Former. Then, we concatenate the processed text features and image features, together with the action nosied by a DDPM scheduler with 100 time-step. The multimodal inputs then pass through a causal Transformer and predict the added noise, which will execute on the robot arm after a series of post-processing. The Transformer is a from scratched Llama2-type model with 12 self-attention blocks. The hidden size is set to 768. All the modules mentioned are trained except the text encoder of clip. In summary, we have 334M parameters in total and 221M trained. This is pioneering to get this performance with such a small-sized model.

\section{Objects and Tasks Description}
\label{sec:tasks}

The object in our experiments is usually small. The banana is 3cm width and 16cm long. The green block is 2cm width and 4cm long. The kiwifruit is very light with 3cm width and about 4.5cm long. The tiny block is only 0.5cm width and 4cm long. {\textit please notice the images (224*224) in Figure~\ref{fig:pickup}, Figure~\ref{fig:task_longhorizon}, Figure~\ref{fig:complex1} are the input for our model.} We demonstrate the pick and place task in Figure~\ref{fig:pickup}. In particular, we use the teaching strategy for pick and place tasks to collect data in our experiments. Therefore, we can collect data without remote control or mouse. We further use the remote control to collect more complex data. We demonstrate the complex task and long-horizon tasks as follows,

{\bf Pick up the bowl and pour the balls into the box}. This task requires to pick up the bowl and pour the balls in the bowl into the box. The bowl is much larger than the block in our experiments. Therefore, we find the model is able to achieve better performance for picking up the bowl. Meanwhile, compared to pick and place tasks, the box of this task is fixed. Thus, it is slightly easier.

{\bf Open the box and pick up the green block}. In this task, we place the box below the gripper with some variations. The variation of this task is less than the pick up tasks.

{\bf Stack the bowls}. In this task, we stack the two bowls in the side into the center. For this task, we evaluate it with 20 trials with bowl position variation.

{\bf Pick up the banana and move it into the pen container}. This task aims to pick up the banana and insert into the small pen container. The banana is aroud 3cm width and 16cm long, while the pen container is 65cm wide. Therefore, it is very difficult to put the banana into the pen container. For this task, we evaluate it with 20 trials with different banana places.

{\bf Pick up the banana and move it into the pen container}. This task aims to pick up the banana and insert into the small pen container. The banana is aroud 3cm width and 16cm long, while the pen container is 65cm wide. Therefore, it is very difficult to put the banana into the pen container. For this task, we evaluate it with 20 trials with different banana places.

{\bf Pick up the mark cup and pouring coffee beans from cup into the bowl}. This task aims to pick up the mark cup, and then pouring the coffee beans from the cup into the bowl on top of the box. The mark cup is 200g. Therefore, it is very difficult to pick up it successsfully.

\section{Analysis and Discussion}
\label{sec:analysis}

\subsection{Comparisons and Analysis with SoTA}
\label{sec:analysis:long}

In our experiments, we find OpenVLA~\cite{kim2024openvla} achieves good pick-up performance compared Octo~\cite{team2024octo}. However, when the task requires learning rotation operation, Octo is able to achieve better performance, e.g., Open the box. We think it is because Octo predicts continuous actions compared to OpenVLA, and the diffusion policy is not sensitive to action normalizations. Differently, OpenVLA requires to discretize the actions dependent on the action statistics. In our experiments, we calculate the statistics from the 10-shot training samples (all tasks) for OpenVLA, Octo, and DiT policy. It is difficult to obtain suitable statistics for Discretization values. However, diffusion policy is able to predict the continuous actions.


\subsection{Zero-shot generalization}

Figure~\ref{fig:zero_shot_real} presents the proposed method is able to grasp the object, while OpenVLA and Octo fails. We observe OpenVLA and Octo fail to rightly approach the right grasp position. This experiment demonstrates the proposed Diffusion Transformer structure achieves more robust policy learning compared to discretized actions or diffusion action head. The denoising transformer model has built a better mapping from the image observation to corresponding action chunks.


\begin{figure}
    \centering
    \includegraphics[width=0.9\linewidth]{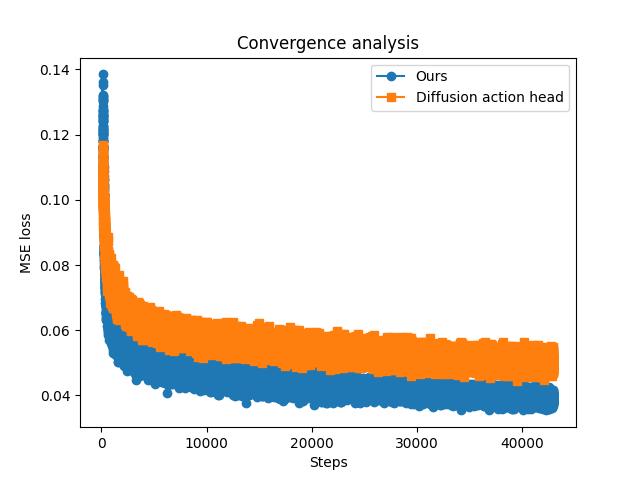}
    \caption{Convergence Analysis on OXE dataset~\cite{brohan2023rt}. The blue line is DiT Policy, and the orange line is Diffusion action head strategy with the same number of parameters.}
    \label{fig:convergence}
\end{figure}

\subsection{Convergence Analysis}
\label{sec:analysis:converge}
Figure~\ref{fig:convergence} presents the convergence between the baseline method and DiT Policy. We can find that the DiT policy achieves clear faster convergence compared to the diffusion action head strategy. We think this also demonstrates the scalibity of DiT Policy.

\subsection{Analysis in Calvin}
As illustrated in the main paper, we use a common learning rate scheduler to decay the learning rate in the experiments in Calvin, rather than a fixed learning rate of 0.0001 in our pre-training stage. We demonstrate that this can slightly improve the performance in Table~\ref{tab:calvin_lr}.


%


%


 \begin{table}[t]
 \setlength\tabcolsep{4pt}
 \caption{The Ablation of learning rate scheduler on Calvin Benchmark.}

 \centering
  \label{tab:calvin_lr}
  \small
  
  \begin{tabular}{c|c|c|c|c|c|c}
   \toprule
     {\bf strategy } & \multicolumn{6}{|c}{No. Instructions in a Row (1000 chains)} \\ 
    \hline
     w lr decay & {\bf 94.5}\%  & {\bf 82.5}\%  & {\bf 72.8}\% & {\bf 61.3}\% & {\bf 50.0}\% & {\bf 3.61} \\
     w/o lr decay & 91.8\%  & 80.0\%  & 68.0\% & 56.9\% & 45.9\%  & 3.43 \\
   \bottomrule
  \end{tabular}

\end{table}

\section{Comparisons on More Diffusion Strategies}
\label{sec:diffusion_baselines}
In our experiments, we implement a strong baseline based on the core idea of diffusion action head. However, it is slightly different from Octo~\cite{team2024octo} when we predict action chunks. Octo~\cite{team2024octo} flattens the action chunks into a single vector with a single embedding. For example, if it requires to predict 8 actions, they predict a $8*7=56$ vector. Unlike the Octo-style diffusion action head, we implement a shared diffusion action head as illustrated in Figure 1 in the main paper, which is more effective. Specially, it is more similar to the Diffussion loss~\cite{li2024autoregressive} and Diffusion Force~\cite{chen2024diffusion}. We further devise different diffusion heads, including Unet1D, Transformer on Calvin without pretraining.

Table~\ref{tab:maniskill_main} presents the comparisons about different baseline methods. We can find Diffusion Head with action chunks~\cite{team2024octo} achieves poor performance on Maniskill2. This indicates that the proposed diffusion action head baseline is more effective compared to the action chunks design~\cite{team2024octo}. Meanwhile, when we increase the blocks of the action head in our baseline, we do not find clean improvement compared to the three layer MLP. Moreoever, the proposed DiT policy achieves clean better performance on Maniskill2, especially on complex YCB tasks.

Table~\ref{tab:calvin_baselines} shows that the proposed Dit Policy achives the better generalization on Calvin (ABC$\rightarrow$D), compared to other diffusion head strategies~\cite{bharadhwaj2024roboagent,zhao2023learning,chi2023diffusion}.

\begin{table*}[t]
\caption{Comparision with more diffusion baseline methods on Maniskill2 under success rate. SingleYCB indicates PickSingleYCB, ClutterYCB indicates PickClutterYCB, SingleEGAD indicates PickSingleEGAD. Disc ActionHead indicates Discretized Action Head strategy~\cite{brohan2022rt}, while Diff ActionHead w/ action chunks is totally similar to Octo strategy~\cite{team2024octo}. Diffusion Action Head is our baseline. Diffusion Action Head w/ 8 MLP blocks is our baseline with 8 blocks MLP action head.}
\label{tab:maniskill_main}
\small
\begin{center}
\begin{tabular}{l|c|ccccc}
\toprule
{\bf Method}  & {\bf All} & {\bf PickCube} & {\bf StackCube} & {\bf SingleYCB} & {\bf ClutterYCB} & {\bf SingleEGAD} \\
\hline 
Disc ActionHead & 30.2\% & 41.0\% & 33.0\% & 22.0\% & 1.0\% & 54.0\% \\
Diffusion Head w/ single token action chunks~\cite{team2024octo} & 36.4\% &  68.0\% &  51.0\% &  26.0\% &  4.0\% &  33.0\% \\
Diffusion Action Head& 58.6\% &  86.0\% &  76.0\% &  37.0\% &  24.0\% &  70.0\% \\

Diffusion Action Head w/ 8 MLP blocks & 57.6\% & {\bf 87.0\%} &  73.6\% &  58.3\% &  20.8\% &  53.1\% \\


DiT Policy(ours) & {\bf 65.8\%} &  79.0\% &  {\bf 80.0\%} &  {\bf 62.0\%} &  {\bf 36.0\%} &  {\bf 72.0\%}  \\

   \bottomrule
\end{tabular}
\end{center}

\end{table*}

\begin{table}
\setlength\tabcolsep{2pt}
\caption{More action designs w/o pretraining on Calvin (ABC$\rightarrow$D). 
}

\centering
\label{tab:calvin_baselines}
\small

\begin{tabular}{c|c|c|c|c|c|c}
\toprule
 {\bf Methods } & \multicolumn{6}{|c}{No. Instructions in a Row (1000 chains)} \\ 
\hline
\hline 
Unet1D head~\cite{chi2023diffusion} & 76.8\% & 46.5\% & 28.8\% & 18.5\% & 10.0\% & 1.80 \\
Transformer head~\cite{chi2023diffusion} & 75.8\% &  44.8\% & 26.5\% &16.5\% & 8.0\% & 1.72 \\
8 layers MLP head & 69.8\% & 42.5\% & 26.3\% & 16.8\% & 11.0\% & 1.66\\
3 layers MLP head & 75.5\%  & 44.8\%  & 25.0\% & 15.0\% & 7.5\% & 1.68 \\
\hline
Single token act chunks & 56.5\% & 18.3\% & 6.0\% & 2.8\% & 0.8\% & 0.84 \\
Ours & {\bf 89.5\%}  & {\bf 63.3\%}  & {\bf 39.8\%} & {\bf 27.3\%} & {\bf 18.5\%} & {\bf 2.38} \\

   \bottomrule
\end{tabular}
\vspace{-4mm}
\end{table}

\section{Additional Details about Pretraining Data}
\label{sec:data}

We choose 15 large datasets from Open X-Embodiment~\cite{padalkar2023open} as illustrated in Table~\ref{tab:data_mix}. We mainly follow~\cite {team2024octo,kim2024openvla} to set the weights. Following~\cite{kim2024openvla}, we further resize the image to the size of 224.

\begin{table}[h]
\caption{The training dataset mixture}
\small

  \centering
  \begin{tabular}{c|c|c}
  \toprule
    Fractal~\cite{brohan2022rt} & DobbE~\cite{shafiullah2023bringing} & Droid~\cite{khazatsky2024droid} \\
    
     \hline
    16.15   & 1.94 &  13.69 \\
    \hline\hline
    Robo Set~\cite{kumar2024robohive} & Viola~\cite{zhu2023viola}  & Kuka~\cite{kalashnikov2018qtoptscalabledeepreinforcement} \\
    \hline
    2.99 & 1.30 & 17.47 \\
    \hline\hline
    BridgeV2~\cite{walke2023bridgedata} & NYU Franka ~\cite{cui2022play} & Furniture Bench~\cite{heo2023furniturebench} \\
    \hline
    21.86 & 1.14 & 6.73 \\
    \hline\hline
    StanfordHydra~\cite{belkhale2023hydra} & DLR EDAN ~\cite{quere2020shared} & BerkeleyFanuc ~\cite{zhu2023fanuc} \\
\hline
 6.11 &0.08 & 1.07  \\
    \hline\hline
    Jaco~\cite{dass2023jacoplay} & LanguageTable~\cite{lynch2023interactive} & toto~\cite{zhou2023train} \\
    \hline
     0.67 & 6.01 & 2.78 \\

        \bottomrule
  \end{tabular}
  
  \label{tab:data_mix}
  
\end{table}


\begin{figure*}
    \centering
    \includegraphics[width=0.95\linewidth]{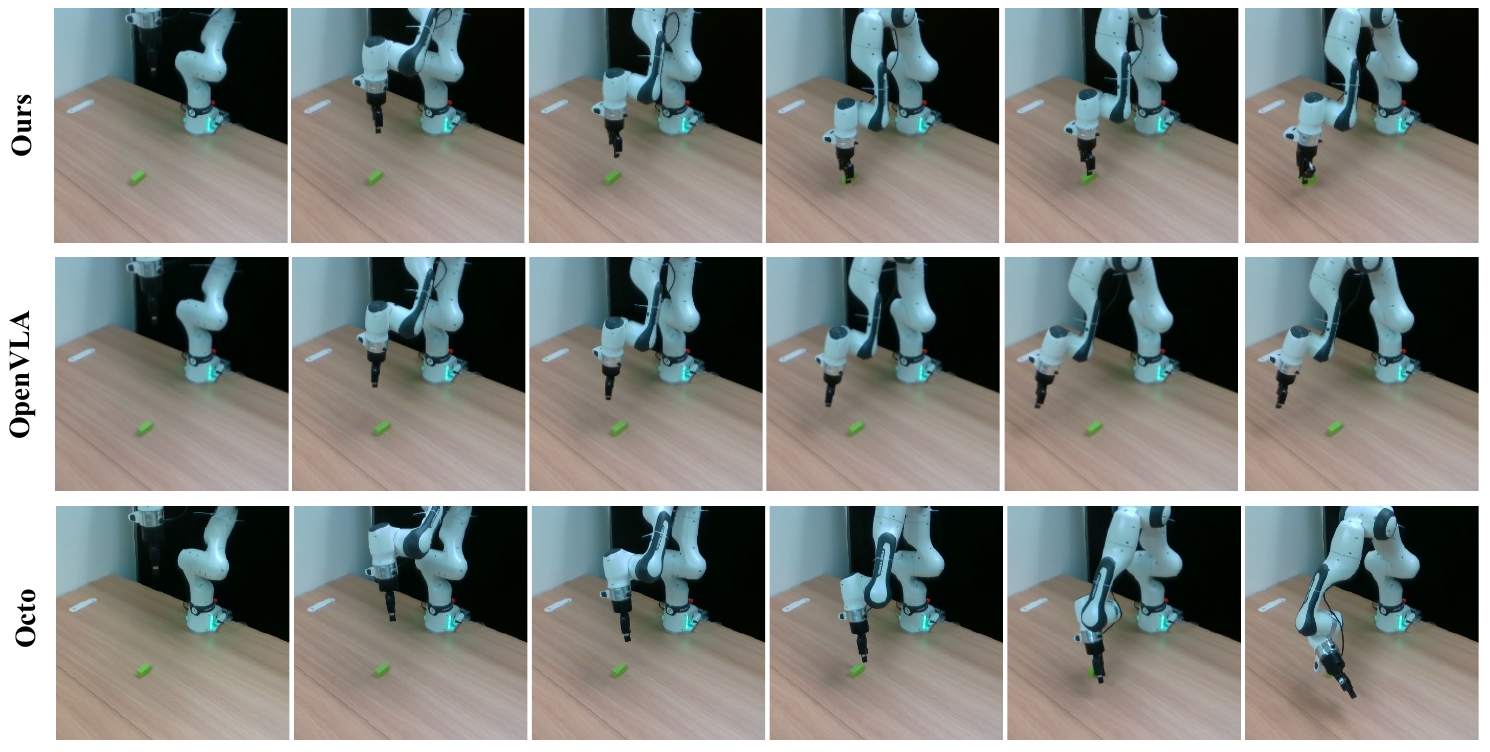}
    \caption{Visualization of zero-shot genearalizaton of different models. The first raw is Diffusion Transformer Policy(ours). The second raw of demonstration is OpenVLA~\cite{kim2024openvla}, the third raw is Octo~\cite{team2024octo}. Our Model can complete the tasks successfully while both of the others fails.}
    \label{fig:zero_shot_real}
\end{figure*}

\begin{figure*}
    \centering
    \includegraphics[width=1.\linewidth]{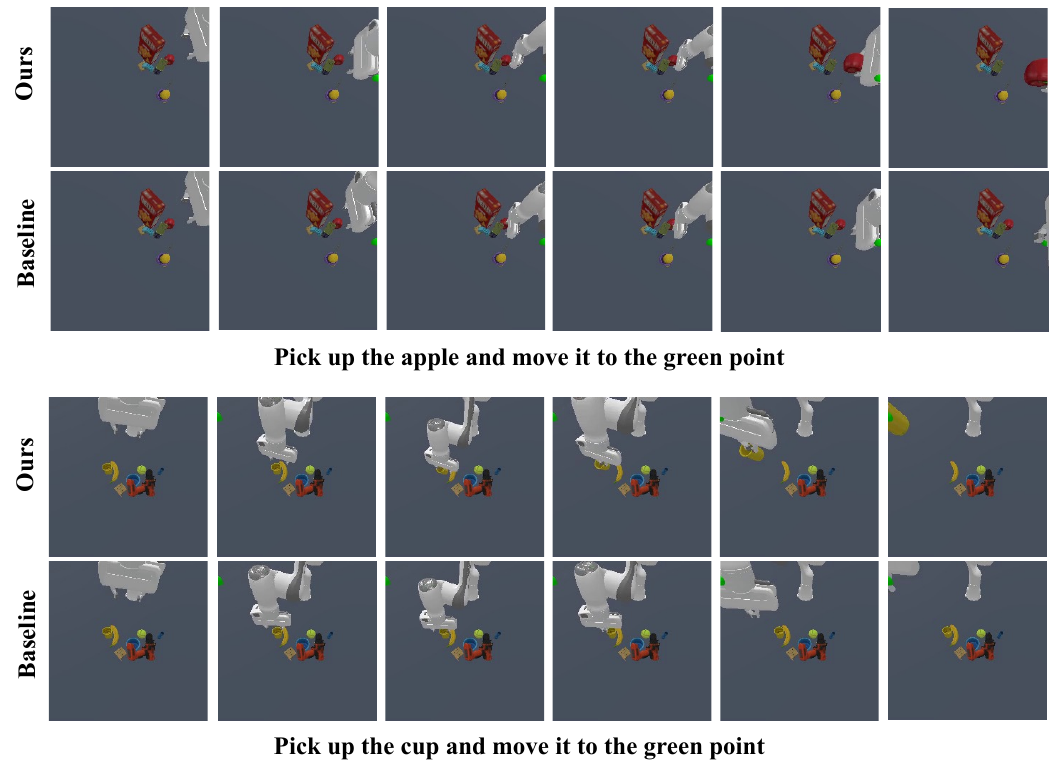}
    \caption{Visualized comparison between Diffusion Transformer Policy and Diffusion Action Head baseline on Maniskill2 (PickClutterYCB). The first raw is Diffusion Tranformer Policy, while the second raw is the baseline method with Diffusion Action Head. }
    \label{fig:vis_comp}
\end{figure*}

\begin{figure*}
    \centering
    \includegraphics[width=1.\linewidth]{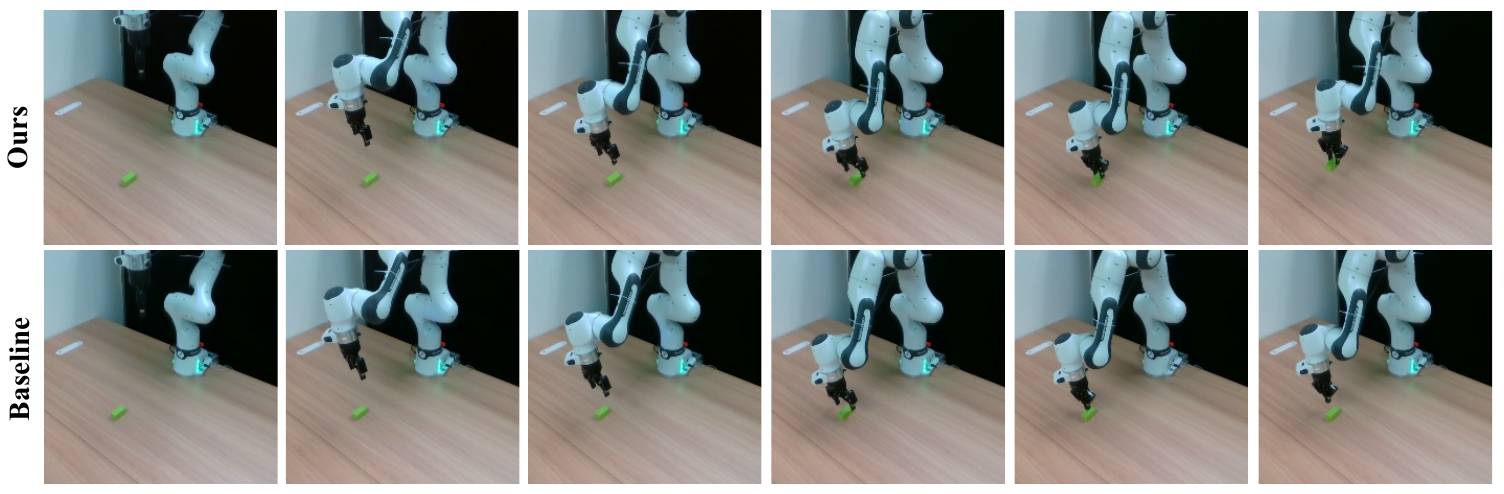}
    \caption{Visualized comparison between Diffusion Transformer Policy and Diffusion Action Head strategy on Real Franka Arm (Pick up the green block). The first raw is DiT Policy, while the second raw is the baseline method with Diffusion Action Head. }
    \label{fig:vis_comp_real}
\end{figure*}





\end{document}